# Anomaly Detection Based on Deep Learning Using Video for Prevention of Industrial Accidents

Satoshi Hashimoto,[1] Yonghoon Ji,[1] Kenichi Kudo,[2] Takayuki Takahashi,[3] and Kazunori Umeda[1]

*Abstract*—This paper proposes an anomaly detection method for the prevention of industrial accidents using machine learning technology. The number of casualties due to industrial accidents is large all over the world.

Currently, a comprehensive system for preventing such accidents is basically done manually. Furthermore, it has not been automated. In order to carry out anomaly detection reliably, a sequential skeleton map of a person is used as training data from skeleton information extracted by OpenPose. Here, long short-term memory-based variational autoencoder (LSTM-VAE) is applied as a deep learning model. It is confirmed that this produces detection results with greater accuracy than those of conventional methods.

## I. INTRODUCTION

The number of casualties due to industrial accidents is large around the world. For example, in Japan, 127,329 people were injured in industrial accidents in 2018 [1]. With increasing numbers of foreign and elderly workers, the manufacturing industry is becoming more diversified. For this reason, the decrement in the number of industrial accidents, which had been a trend in Japan, has slowed. Figure 1 shows the number of industrial accidents in Japan over time.

Once industrial accidents occur, various responsibilities are imposed on the company. Therefore, various measures have been taken to prevent industrial accidents. Risk assessment is one of these measures [2]. Risk assessment generally refers to a series of steps that identify hazards, assess risks, and take corrective action accordingly. Steps of risk assessment are shown in Fig. 2. However, a comprehensive system to prevent industrial accidents, such as risk assessment, is basically done manually. Research on automating risk assessment is scarce and cannot be practically achieved.

On the other hand, several studies have been conducted on detecting anomalies in real-world situations in daily life [3–6]. In general, anomaly detection is a data-mining method that defines the degree of anomaly from a normal distribution using an unsupervised machine learning method. The steps of anomaly detection are shown in Fig. 3. Anomaly detection is expected to be used in various fields, for example, detecting criminal behavior using video acquired from security cameras or falling accidents in daily life by recognizing motion.

There are various methods to detect anomalies. Principal component analysis (PCA) is a method that learns a subspace representing normal data and calculates the degree of anomaly from the distance of the normal distribution. A Gaussian Mixture Model (GMM) is another method for detecting anomalies by estimating multiple normal distributions by using the maximum likelihood method.

In recent years, methods using deep learning technology have appeared in the anomaly detection field. Autoencoder (AE) [7] is a neural network of the encoder-decoder type that encodes high-dimensional data, such as images, into latent space and decodes images from the acquired features. In addition, variational autoencoder (VAE) [8] incorporates Bayesian theory into the autoencoder. By assuming a normal distribution in the latent space, it is possible to generate images with higher accuracy.

However, when dealing with high-dimensional data such as images and videos, conventional methods such as PCA and GMM have problems with the curse of dimensionality, in that the interclass distance and the intraclass distance narrow in high-dimensional space. Therefore, it is difficult to detect anomalies with high accuracy. For high-dimensional data input, deep learning methods, such as AE and VAE, are often used. However, these frameworks make assumptions that those input are independent and identically distributed (i.i.d.) data. Therefore, these frameworks are not suitable because they cannot handle time series correlations when detecting anomaly motions.

Therefore, in this paper, we set two goals based on the above background. One is to build a comprehensive automatic monitoring system that prevents industrial accidents using anomaly detection methods.

The second is to enable anomaly detection with higher accuracy than by conventional methods by using a deep learning method that can handle time series information.

Because of [2] and the occurrence situation of industrial accidents in Japan, we consider that most causes of industrial accidents occur when workers deviate from the rules established in the factory.

In order to prevent industrial accidents, we focus in this paper on detecting abnormal behavior using video that can be acquired from cameras installed in factories. In addition, in order to recognize such behavior, we will perform anomaly detection using long short-term memory-based variational autoencoder (LSTM-VAE) [9], which can consider time-series input. As a concrete task, we take up the anomaly detection of multiple behavior patterns that imitate work in the factory and confirm the accuracy of our method by the verification experiment.

In this paper, we first introduce a framework for unsupervised anomaly detection using deep learning. Next, we introduce verification experiments. Finally, we verify the effectiveness of our method.

1 The Course of Precision Engineering, School of Science and Engineering, Chuo University, Tokyo, Japan
(corresponding author's e-mail: hashimoto@sensor.mech.chuo-u.ac.jp).
2 Chuo University, Tokyo, Japan
3 Prima Meat Packers, Ltd., Tokyo, Japan

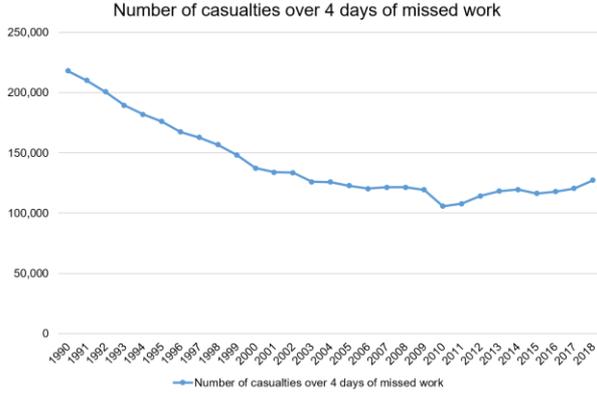

Fig. 1 The number of industrial accidents in Japan

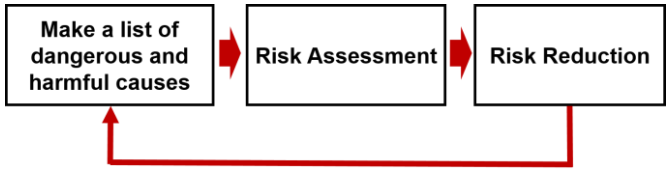

Fig. 2 The steps of risk assessment

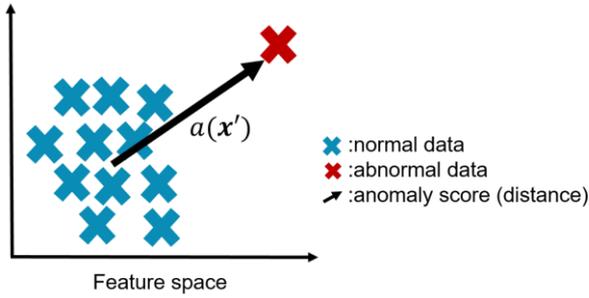

Fig. 3 Anomaly detection

## II. METHOD

### A. Method Overview

The steps of the proposed method are shown in Fig. 4. First, the behavior of a worker is captured on video using a camera, and OpenPose is used to acquire the worker's skeleton information. OpenPose is a library that implements the skeleton estimation method of Cao et al. [10], with which it is possible to detect 25 skeleton points of persons from one image by using a convolutional neural network (CNN).

Next, the acquired skeleton image is converted by bilinear interpolation. Here, a time series subsequence video having a length of time step $T$ is created using a sliding window method. We named it a sequential skeleton map.

Next, training is performed by LSTM-VAE using this map as input. When performing training, a model is trained using a data set that contains normal behavior prepared in advance.

Finally, anomaly detection is performed by using the model learned by the proposed method. The degree of anomaly is calculated using the value of the error function between the input sequential skeleton map and the output LSTM-VAE.

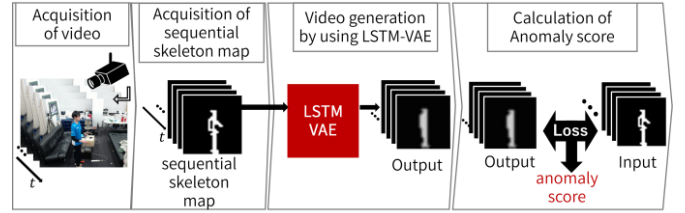

Fig. 4 The steps of the proposed method

### B. Acquisition of Skeleton Information Using OpenPose

We use OpenPose to acquire a person's skeleton information for the captured input video. Next, we apply preprocessing to the acquired skeleton images. The steps of preprocessing are shown in Fig. 5.

First, we cut skeleton images to a size of 480 × 480, centering on the coordinates of the neck. Next, the resolution is converted to 28 × 28. Here, a bilinear method is used as the interpolation method.

Finally, binarization is performed. A sequential skeleton map with length $T$ is created using the sliding window method for the acquired 28 × 28 skeleton images. The width between adjacent sequential skeleton maps in sliding window is 1. This will be the input to LSTM-VAE.

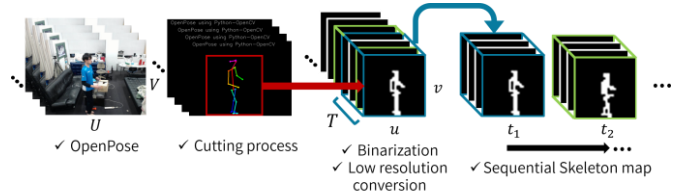

Fig. 5 Preprocessing flow

### C. Anomaly Detection Using LSTM-VAE

In this paper, we use an unsupervised learning method that can train only normal data because it is difficult to acquire abnormal data. In addition, by using LSTM-VAE, which is a deep learning generative model that can consider time series, output similar to the input sequential skeleton map is generated, and anomaly detection is performed from the error score between the input and output.

The architecture of LSTM-VAE in this method is shown in Fig. 6. LSTM-VAE is VAE with an LSTM layer that enables input to consider time series information. Long short-term memory (LSTM) [11] is one of the time series neural networks that solved the vanishing gradient problem in a recurrent neural network (RNN) by using the concept of human memory.

In this method, features of each frame's image of sequential skeleton maps are extracted using CNN. The features acquired by CNN are input to the LSTM layer with the length of time step $T$, and time series information is modeled. The dimensionality of the output from LSTM is reduced and compressed to a latent space $N(\mathbf{0}, \mathbf{I})$ assuming a normal distribution. Using this latent space, we can generate output similar to the input by decoding a sequential skeleton map. In addition, the error function used for learning is shown in equation (1).

Here, $\mathbf{x}'$ and $\mathbf{y}'$ are input and output, respectively. Its dimension is represented by D. $\boldsymbol{\sigma}$ is the variance vector, and $\boldsymbol{\mu}$ is the mean vector. $J$ is the dimension of the latent variable. $\mathbb{E}$ is the expectation value.

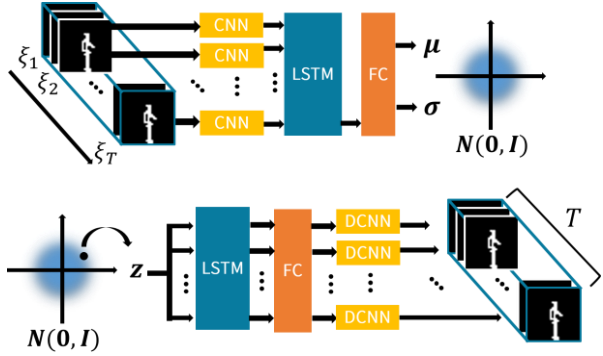

Fig. 6 The architecture of LSTM-VAE

$$L(\mathbf{x}') = \frac{1}{2}\sum_{j=1}^{J}\left(1 + ln(\sigma_j^2) - \mu_j^2 - (\sigma_j^2)\right) + \mathbb{E}\left[\sum_{i=1}^{D}(x_i' ln y_i' + (1 - x_i')ln(1 - y_i'))\right] \quad (1)$$

## III. EXPERIMENTS

### A. Data Set

Verification experiments were conducted to confirm the effectiveness of the proposed method and to compare its accuracy with that of the conventional method using AE and VAE.

First, in order to acquire a sequential skeleton map used for normal training data, we made a data set that included several patterns of actions that imitate work in the factory. The list of actions is shown in Table 1. Training data, which contain normal actions, are used for LSTM-VAE training. Then, the model's performance is evaluated by a test data set that includes abnormal action. These actions are made with reference to the deviant actions that can occur in the actual factory. Then, using a monocular surveillance camera installed at a height of 2.7 m from the ground, the situation is recorded at an imaging speed of 30 fps with a resolution of 640 × 480. From the video recorded in this way, we acquired 4,618 frames of skeleton images that contain only normal motions. The length of this data is about 3 minutes.

Then, a sequential skeleton map was created with the number of time steps $T$ = 30, and 4,589 sequential skeleton maps were acquired and used for training. Similarly, we acquired skeleton images of 4,898 frames including abnormal motions used for testing. The length of this data is about 3 minutes. We acquired 4,758 sequential skeleton maps in the same way as described above.

TABLE 1 The list of actions

| Work | Normal action (for training data set) | Abnormal action (for test data set) |
|---|---|---|
| 1 | Carry the chair with both hands | Carry the chair with one hand |
| 2 | Cut the paper with both hands | Cut the paper with one hand |
| 3 | Discard trash in front of the box | Discard trash by throwing |
| 4 | Carry the box with both hands | Carry the box with one hand |
| 5 | - | Dig in the trash box |
| 6 | - | Work while running |

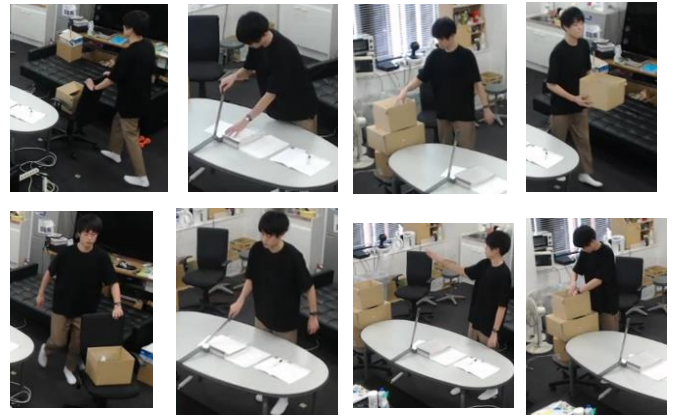

Fig. 7 Examples from our data set

### B. Training

We performed LSTM-VAE learning using 4,589 sequential skeleton maps including only normal actions.

In addition, we also trained AE and VAE as a comparison method. Note that the input to AE and VAE includes skeleton images of 4,618 frames.

Table 2 shows the parameters of training list for each method. These hyperparameters are used by evaluating the value that seems to be optimal by quantitative evaluation using the area under the receiver operating characteristic (AUROC) value described later. We used Ubuntu 16.04 LTS for the OS and a GPU of NVIDIA GTX 1080 Ti (11 GB) for training.

### C. Evaluation

For evaluation, we used AUROC, which is the size of the area enclosed by the ROC curve. ROC is an index for evaluating the accuracy of a model. The ROC curve is a general evaluation index of anomaly detection methods; its horizontal axis is the false positive rate (FPR), and the vertical axis is the true positive rate (TPR). ROC curves are plotted by changing the threshold that separates normal from abnormal. We show that the performance is better because the model with the ROC curve on the upper left has a larger AUROC.

In addition, in order to evaluate overfitting, it is evaluated whether the optimal training can be done by plotting the transition of the loss function. Note that batch normalization (BN) [12] is performed on each CNN layer in order to prevent overfitting.

TABLE 2 Parameters of the training phase

| Parameters | LSTM-VAE | AE | VAE |
|---|---|---|---|
| Epochs | 30 | 30 | 30 |
| $J$ | 16 | 16 | 16 |
| $T$ | 30 | - | - |
| Latent dimension of LSTM | 300 | - | - |
| Batch size | 16 | 32 | 32 |

*D. Experimental Results*

The ROC curve is shown in Fig. 8, and the transition of the anomaly with the anomaly score on the vertical axis and the frame on the horizontal axis is shown in Fig. 9. In the graph of the anomaly, the colored ranges are the frame in which the abnormal action exists. Table 3 shows the AUROC value of the model.

In the proposed method, it was possible to detect abnormal action patterns with high accuracy. In addition, we have achieved anomaly detection with higher accuracy as compared to AE and VAE, which are the conventional methods. As compared to the conventional methods, the ROC curve and the AUC value both show better accuracy. An example of the sequential skeleton map generated by LSTM-VAE is shown in Fig. 10. Here, the upper part of each image is the input, and the lower part is the output. The distribution of the latent space of LSTM-VAE is shown in Fig. 11. To display the latent space, t-SNE [13] is used to compress the dimensions for the latent space in two dimensions.

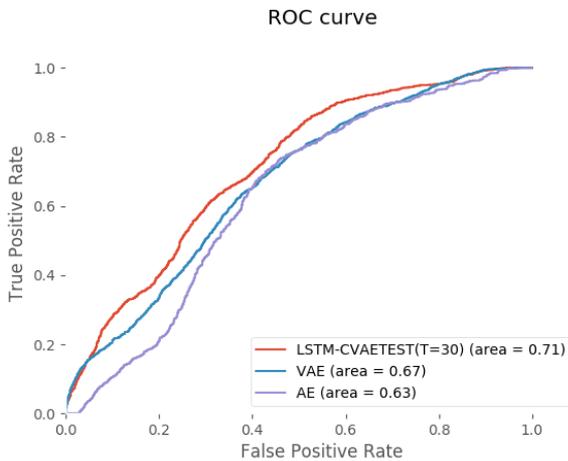

Fig. 8 ROC curve

TABLE 3 AUROC values

|  | AUROC |
|---|---|
| LSTM-VAE | 0.71 |
| AE | 0.67 |
| VAE | 0.63 |

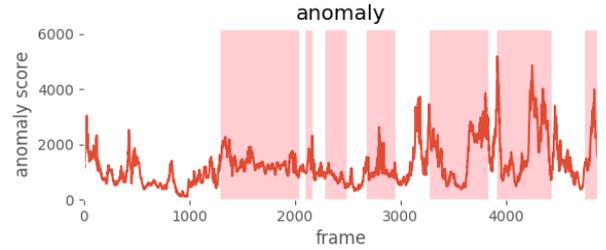

Fig. 9 The transition of the anomaly score

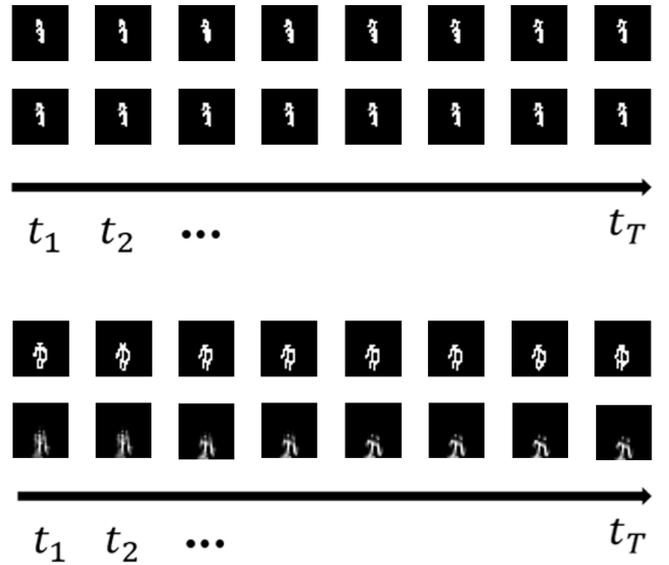

Fig. 10 Sequential skeleton map generated by LSTM-VAE

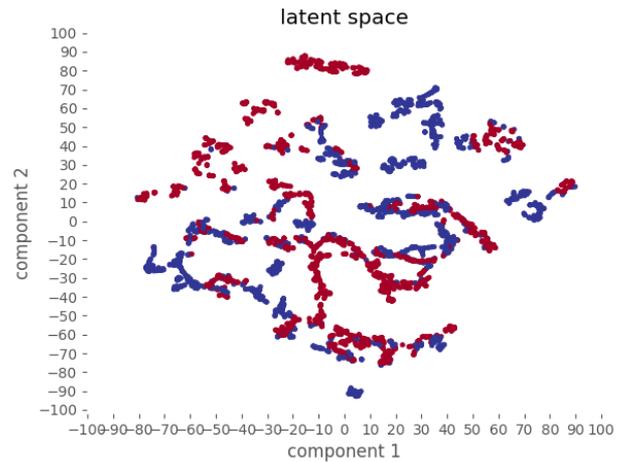

Fig. 11 The distribution of the latent space of LSTM-VAE

## IV. Conclusion

In this study, we developed an anomaly detection method for the prevention of industrial accidents using deep learning. In order to carry out anomaly detection reliably, we used a sequential skeleton map as LSTM-VAE input. We confirmed that detection results with higher accuracy are produced as compared with the conventional methods.

In the future, we will develop a novel method by using generative adversarial networks (GAN), which can generate skeleton maps with high accuracy as compared to a conventional method. We are also considering making the input to the model a raw video. In addition, in order to realize a comprehensive system of preventing industrial accidents, we have examined methods of environmental recognition, such as surrounding objects and positional relationships. We plan to develop such environmental recognition methods.